\documentclass{article}

\PassOptionsToPackage{numbers, compress}{natbib}
\usepackage[final]{nips_2016}

\usepackage[utf8]{inputenc} % allow utf-8 input
\usepackage[T1]{fontenc}    % use 8-bit T1 fonts
\usepackage{hyperref}       % hyperlinks
\usepackage{url}            % simple URL typesetting
\usepackage{booktabs}       % professional-quality tables
\usepackage{amsfonts}       % blackboard math symbols
\usepackage{nicefrac}       % compact symbols for 1/2, etc.
\usepackage{microtype}      % microtypography
\usepackage{bbm}
\usepackage{bm}
\usepackage{amsmath}
\usepackage[pdftex]{graphicx}
\usepackage{subfig}

\title{Bayesian optimization with shape constraints}

\author{
  Michael Jauch \\ %\thanks{Use footnote for providing further
  Department of Statistical Science\\
  Duke University\\
  \texttt{michael.jauch@duke.edu} \\
  \And
  V\'ictor Pe\~na\\
  Department of Statistical Science\\
  Duke University\\
  \texttt{vp58@stat.duke.edu} \\
}

\begin{document}
% \nipsfinalcopy is no longer used

\maketitle

\begin{abstract}
    In typical applications of Bayesian optimization, minimal assumptions are made about the objective function being optimized. This is true even when researchers have prior information about the shape of the function with respect to one or more argument. We make the case that shape constraints are often appropriate in at least two important application areas of Bayesian optimization: (1) hyperparameter tuning of machine learning algorithms and (2) decision analysis with utility functions. We describe a methodology for incorporating a variety of shape constraints within the usual Bayesian optimization framework and present positive results from simple applications which suggest that Bayesian optimization with shape constraints is a promising topic for further research. 
\end{abstract}

%    In typical applications of Bayesian optimization, minimal assumptions are made about the functions to be optimized. This is the case even when researchers have prior knowledge a In important problem domains, this is at odds with researchers' prior understanding of the functions they want to optimize. Such prior knowledge We consider two function classes which, being expensive to evaluate, benefit from a Bayesian optimization treatment: multiattribute utility functions and generalization error surfaces. For both classes there is often prior knowledge that can be expressed as componentwise shape constraints and can be incorporated into the Bayesian optimization framework. We present a variety of shape constraints and their implementation and evaluate their effectiveness on simple real world examples.

\section{Introduction}

%Don’t worry about abstract yet. But, abstract tends to be focused more precisely on what our contribution is. Be more specific in the abstract.
%For introduction, should probably include a few sentences about Bayesian optimization with citations, but no need to describe what it is. Should straight up say what our contribution is and start discussing the details of that. We can add more context and selling later. 
%There are a few examples where people model shape of objective function, but it’s much different and related to more complicated scenarios than typical BO. See freeze-thaw BO and the Klein paper. 

%%%% Stuff below is OLD %%%%

% An advantage of the Bayesian approach is that it allows one to incorporate prior information about the objective function. 
% From early on, the standard choice has been the Gaussian process. Many subsequent developments are built upon this foundation. We stick with it. 
% The use of prior knowledge has been somewhat limited. [Usually just comes through the choice of covariance, which determines smoothness. Other approaches include embedding within a hierarchical model to share information across problems.] Attention has focused on more general priors. 
% We propose that for some function classes, more is known than is typically captured by the prior. 

%%%% Stuff above is OLD %%%% 

Prior knowledge about an unknown function often pertains to the shape of the function with respect to an argument. Consider growth curves in biology: excluding rare incidents, a child's height grows monotonically as a function of his or her age. Incorporating shape constraints such as monotonicity, convexity, or concavity is a well-established topic in function estimation. Such constraints, when warranted, can lead to more efficient and stable function estimates compatible with our prior understanding.

In this paper, we explore the possibility of introducing shape constraints on objective functions in Bayesian optimization. We argue that such constraints are often appropriate in two common application areas: (1) hyperparameter tuning of machine learning algorithms and (2) decision analysis with utility functions. We present an example wherein we optimize the hyperparameters of a support vector machine (SVM) \citep{Cortes:1995:SN:218919.218929} and a problem described in \citet{muller} in which we find the optimal sample size for a binomial experiment. %The application areas explored here are not the only ones for which shape constraints are likely to prove useful, and such constraints are likely to bring greater benefits in more complex problems. Our experiments are meant to illustrate the rationale and application of shape constraints and provide initial evidence of the advantages. 

Since Gaussian processes (GPs) are standard in Bayesian optimization, it would behoove us to stay within that framework. Recent work has looked at ways in which GPs can accomodate shape constraints \citep{aki_monotone}, \citep{Lin23022014}, \citep{xiaojing}. One approach is based on the convenient property that the partial derivative processes of a GP (with a sufficiently smooth covariance function) are jointly Gaussian with the original process \citep{Rasmussen:2005:GPM:1162254}. Thus, one can obtain approximate shape constraints on the original process by imposing conditions on the partial derivatives at a sufficient number of points. We apply this work and expand upon it by introducing a quasiconvexity constraint which is particularly well-suited for applications involving tuning regularization hyperparameters.

%We now offer a heuristic explanation of how shape constraints can improve Bayesian optimization procedures, beyond simply reducing posterior variance. 
The most common prior information used in Bayesian optimization relates to the smoothness of the objective function and is included via the covariance function. %[Shape constraints improve sample efficiency the same way that accurate prior information about smoothness does: by better trading off exploration and exploitation during the search process.] 
While prior knowledge of a function's smoothness allows one to guess what is happening in a neighborhood of an observation, a shape constraint allows one to extrapolate from even a small number of observations to draw conclusions about what is happening globally. Given that certain types of global shape information allow for the design of efficient global optimization algorithms \citep{Boyd:2004:CO:993483}, it is not hard to imagine that such information can also improve the efficiency of Bayesian optimization procedures. Improvements may extend to cases when the objective function only approximately satisfies the shape constraint, or more generally to cases where a convexification of the problem is beneficial.

\subsection{Hyperparameter tuning}
Much of the recent work on Bayesian optimization has focused on the problem of tuning the hyperparameters of machine learning algorithms (see \citet{snoek} for an overview). %In typical applications, a Gaussian process prior is placed on the objective function and little prior information is incorporated other than the smoothness properties implied by the choice of covariance function. 
Recent work by \citet{klein-bayesopt15} and \citet{freezethaw} has taken advantage of the predictable behavior of error or runtime curves to extrapolate what will happen with a larger data set or more training, but there is shape information to be exploited even in more straightforward applications. %If we can adjust our prior distribution to reflect this structure, our algorithms will have improved efficiency over those derived from more standard priors. 

As our primary example, we consider hyperparameters controlling model complexity. %One can think of $\lambda$ in the LASSO objective function: a larger value of $\lambda$ corresponds to a larger penalty on the $\ell_1$ norm of the regression coefficients,  resulting in a lower complexity solution \citep{Tibshirani94regressionshrinkage}. 
Many authors have commented on the characteristic shape of the generalization error curve as a function of such a hyperparameter. See Figure 2.11 of \citet{Hastie09} or Figure 5.3 of \citet{Goodfellow-et-al-2016-Book}. In the latter book, the authors refer to the curves as being "U-shaped." This is not meant to be precise but rather to convey the idea that extreme hyperparameter values, corresponding to very low or very high complexity models, have high generalization error, due to underfitting or overfitting, respectively; the best generalization performance is achieved by a hyperparameter lying somewhere between the extremes. In more mathematical language, we could say these curves are smooth and \textit{quasiconvex} (see section \ref{shapes} for a formal definition of quasiconvexity).

This is useful prior information. Because we can formulate the quasiconvexity constraint in terms of the partial derivatives of the generalization error surface, we can incorporate it into our GP prior using our previous observation about partial derivative processes. We apply this idea in section \ref{svm} to tune the hyperparameters of an SVM.

\subsection{Decision analysis with utility functions}

Another area of potential application is expected utility maximization. In particular, shape constraints are likely to prove useful in optimizing elicited multiattribute utility functions and in Bayesian experimental design. 
\paragraph{Multiattribute utility functions} In some decision problems, the shape of the utility function with respect to an attribute is determined if we accept certain assumptions. For example, utility as a function of monetary rewards is concave and monotonically increasing for a risk-averse agent. Incorporating shape constraints not only improves efficiency by reducing posterior variance, but can correct for systematic elicitation biases that manifest themselves in deviations from the shape constraints dictated by rational choice theory \citep{tversky1975judgment}. 
\paragraph{Bayesian experimental design} In most Bayesian models, the posterior expected utility is a high-dimensional integral which is estimated through Markov Chain Monte Carlo methods. The intensive computation and simulation error involved imply that we can only obtain noisy expected utility estimates at a limited number of design points. It is often the case that the utility as a function of sample size satisfies a convexity or quasiconvexity constraint, reflecting the trade off between uncertainty reduction and sampling costs. In Section~\ref{binomial}, we consider the binomial sample size problem from \citep{muller} with Bayesian optimization and a convexity constraint on the expected utility surface. 

%deisgn variables related to the total sample size s his is another case where the utilities of some design variables are likely to be shape-constrained (especially if they are related to the total sample size) so enforcing them might be useful. [Cite Muller here, mention that they're down with nonparametric shape constraints]% (maybe mention there is a trade-off between uncertainty reduction and sampling costs).

%Bayesian experimental design is another important application area for Bayesian optimization (see \citet{Shahriari} for an overview). In most interesting cases the posterior utility  Maximizing the posterior expected utility often requires costly Monte Carlo . Work by \citet{muller} presents a method in which a curve is fit to Monte Carlo samples of the posterior utility surface. This approach takes advantage of the smoothness of the expected utility surface to share information across design points. The authors prefer parametric models for expected utility surfaces in simple applications, because they allow for the theoretically-motivated shape constraints which are common in this setting; however, for more complex problems, they find that nonparametric models are necessary. Shape constrained Gaussian processes combine the advantages of the parametric and nonparametric approaches. In section \ref{binomial}, we solve the binomial sample size problem from \citep{muller} with Bayesian optimization and a convexity constraint on the expected utility surface. 

\section{Method} 

The shape constraints we are concerned with are \textit{componentwise} shape constraints. Each of the motivating examples involves prior knowledge of the shape of the objective function as a function of one argument, holding all other arguments fixed. We make this notion more precise. Let $f(\bm{x}): \mathbbm{R}^d \to \mathbbm{R}$ be an objective function we wish to minimize over a convex set $\mathcal{A} \subset \mathbbm{R}^d.$  For the $d$-vector $\bm{x},$ let $\bm{x}_{-j}$ be the $(d-1)$-vector omitting the $j^\text{th}$ element of $\bm{x}.$ Then $f(\bm{x}) = f(x_j, \bm{x}_{-j})$ and the function $g_{j,\bm{a}}(x_j) = f(x_j, \bm{a})$ is the component function that arises from fixing the vector $\bm{x}_{-j} = \bm{a}$ and letting $x_j$ vary. The next subsection describes how to implement a variety of shape constraints on component functions.

\subsection{Shape constrained Gaussian processes} \label{shapes}We build most directly on the approach of \citet{xiaojing}. The crucial fact used in that work is that, assuming a sufficiently smooth covariance function, a GP and its partial derivatives form a joint GP. Suppose $f(\bm{x})$ is a GP with covariance function $k(\bm{x}, \bm{x'})$ for $\bm{x}, \bm{x'} \in \mathcal{A}.$ Then
\begin{align} \label{eq:derivative}
    \text{Cov}\left(\frac{\partial^o}{\partial x_j^o} f(\bm{x}), \frac{\partial^{o'}}{\partial x_k'^{o'}} f(\bm{x'})\right) &= \frac{\partial^{o + o'}}{\partial x_j^{o} x_k'^{o'}} k(\bm{x}, \bm{x'})
\end{align}
with $j,k \in \{1, ..., d\}$ and $o,o'$ being the orders of the partial derivatives. We can impose componentwise shape constraints by restricting the partial derivative processes. For example, if we have prior information that $g_{j,\bm{a}}(x_j)$ is monotonically increasing (decreasing) for all $\bm{a},$ we can restrict its partial derivative with respect to $x_j$ to be nonnegative (nonpositive). Similarly, if we want $g_{j,\bm{a}}(x_j)$ to be componentwise convex (concave), we can restrict second partial derivatives to be nonnegative (nonpositive). 

% Paragraph where I define quasiconvexity
An additional and novel constraint which we apply in our hyperparameter tuning application is quasiconvexity (also referred to as unimodality) \citep{Boyd:2004:CO:993483}. We only consider continuous component functions of a single variable, for which it is easy to characterize quasiconvexity. The continuous function $h: \mathbbm{R} \to \mathbbm{R}$ is quasiconvex if and only if (1) $h$ is monotone or (2) there exists some point $c$ such that $h$ is nonincreasing for $t < c$ and $h$ is nondecreasing for $t \geq c.$ Again, this is a condition which can be stated in terms of partial derivatives and is therefore amenable to a GP implementation.

% Clarify that these are approximate and we can't enforce them at a set with a limit point. 
As discussed in \citet{xiaojing}, the partial derivative constraints are only enforced at a discrete set of points. Imposing requirements on the partial derivative at a set having a limit point would eliminate all sample paths of the GP. Realizations from a shape constrained GP only approximately follow the constraint; however, the discrepancy can be made practically irrelevant by enforcing the constraint at sufficiently many points. %This raises the question of how to choose the points at which one enforces the partial derivative constraints economically. We use the algorithm suggested in \citet{xiaojing}.

% Discussion of computation. 
The cases of monotonicity, convexity, or concavity constraints call for sampling high-dimensional truncated Gaussian random variables. We use the methodology introduced in \citet{botev}, which is available in the \textsf{R} package {\tt TruncatedNormal} \citep{botev_package}. For the quasiconvexity case, rejection sampling was efficient enough for the applications presented.

\section{Experiments}
% - Brief description of our Bayes Opt choices: acquisition function (expected improvement), how we are estimating GP parameters (Stan!), how we are maximizing the acquisition function.
In any Bayesian optimization application, one must make choices about the mean and covariance of the GP, the acquisition function, the priors, etc. Before getting to the experiments performed, we fix notation so we can give these details. 
Suppose the data are $n$ observations $y_i = f(\bm{x}_i) + \epsilon_i$ with $i \in \{1, 2, \, ... , \, n\}$ and $\bm{x}_i \in \mathbb{R}^d$. We assume the errors $\epsilon_i$ are independent so that jointly $\bm{y} \sim N_n (\bm{1}_n \mu, \bm{K} + \sigma^2 \bm{I}_n)$, where the $(i,j)$-th entry of $\bm{K}$ is $K_{ij} = k(\bm{x}_i, \bm{x}_j)$. We use a squared-exponential kernel so that $k(\bm{x}, \bm{x}') = \tau^2  \exp \{ -  \sum_{i =1}^d \psi_d (x_{i} - x'_{i})^2 \}$ and place independent Cauchy and half-Cauchy priors on the parameters: $\mu \sim C(0,1)$, $\tau^2 \sim C_+(0,0.25)$, $\sigma^2 \sim C_+(0,0.25)$, and $\psi_d \sim C_+(0,5)$. For ease of computation, we take a \textit{partial likelihood} approach (which amounts to ignoring shape constraints when estimating the parameters, see \cite{xiaojing}) and we don't perform fully Bayesian inference; instead, we plug-in the posterior median found using the \texttt{R} package \texttt{Rstan} \citep{stan}. Finally, the acquisition function we use is expected improvement. We acknowledge that we could likely achieve better results by using a Mat\'ern kernel and performing fully Bayesian inference (as recommended in \cite{snoek}), and by working with an acquisition function that uses the minimum expected first derivative. The merits of both will be explored elsewhere. In our applications, shape constraints are enforced at the observed values of $\bm{x}$ and 100 other locations that are selected by a maximin latin hypercube design using the \textsf{R} package \texttt{lhs} \citep{lhs}. In all the cases considered here, the joint distribution of the observed data and derivatives is multivariate Gaussian with covariance matrices that can be constructed using Equation~\eqref{eq:derivative}.

\subsection{Binomial sample size} \label{binomial}

\begin{figure}[!htbp]
  \centering
  \subfloat[]{\includegraphics[width=0.49\textwidth]{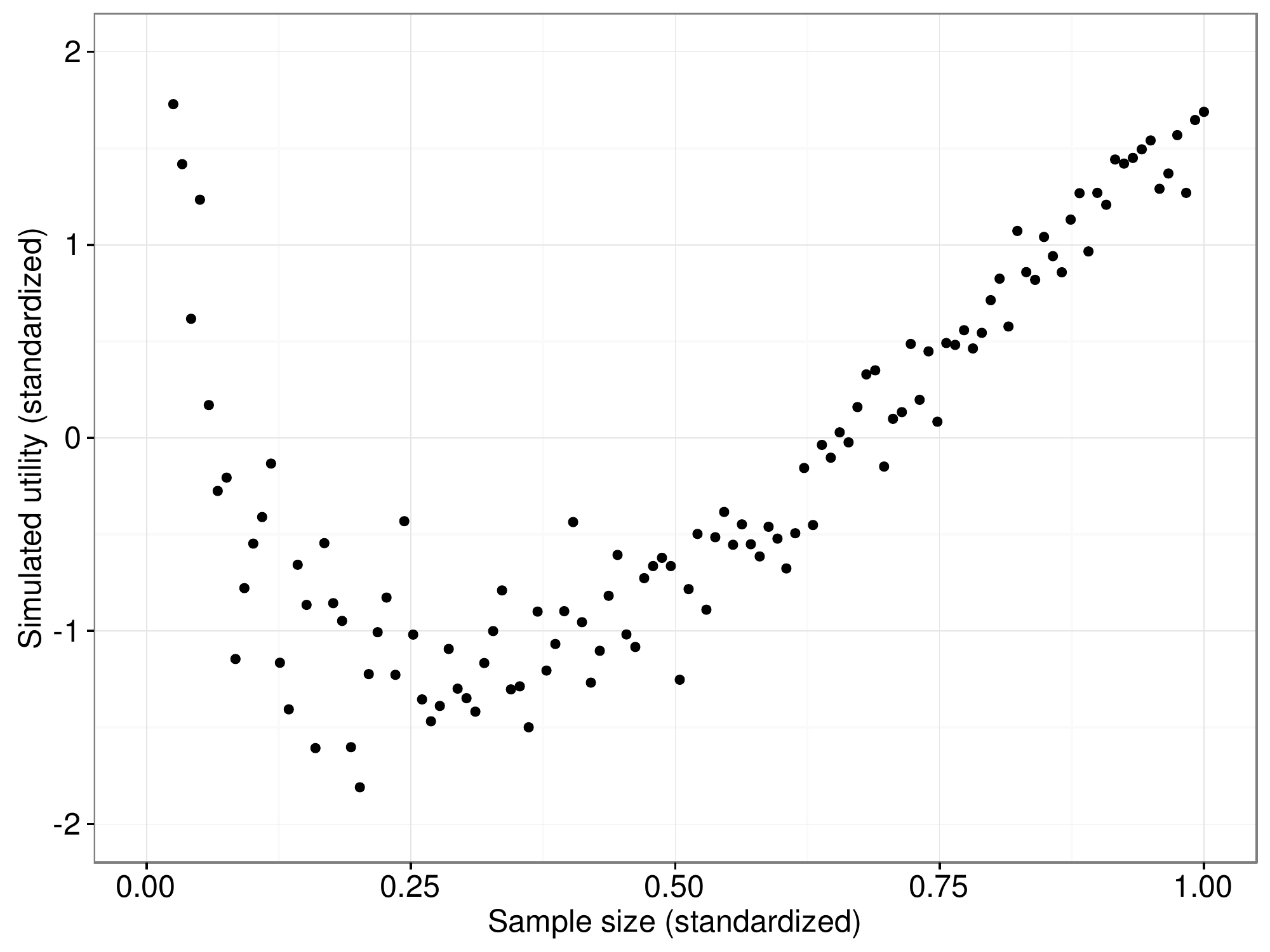}\label{fig:f1}}
  \hfill
  \subfloat[]{\includegraphics[width=0.49\textwidth]{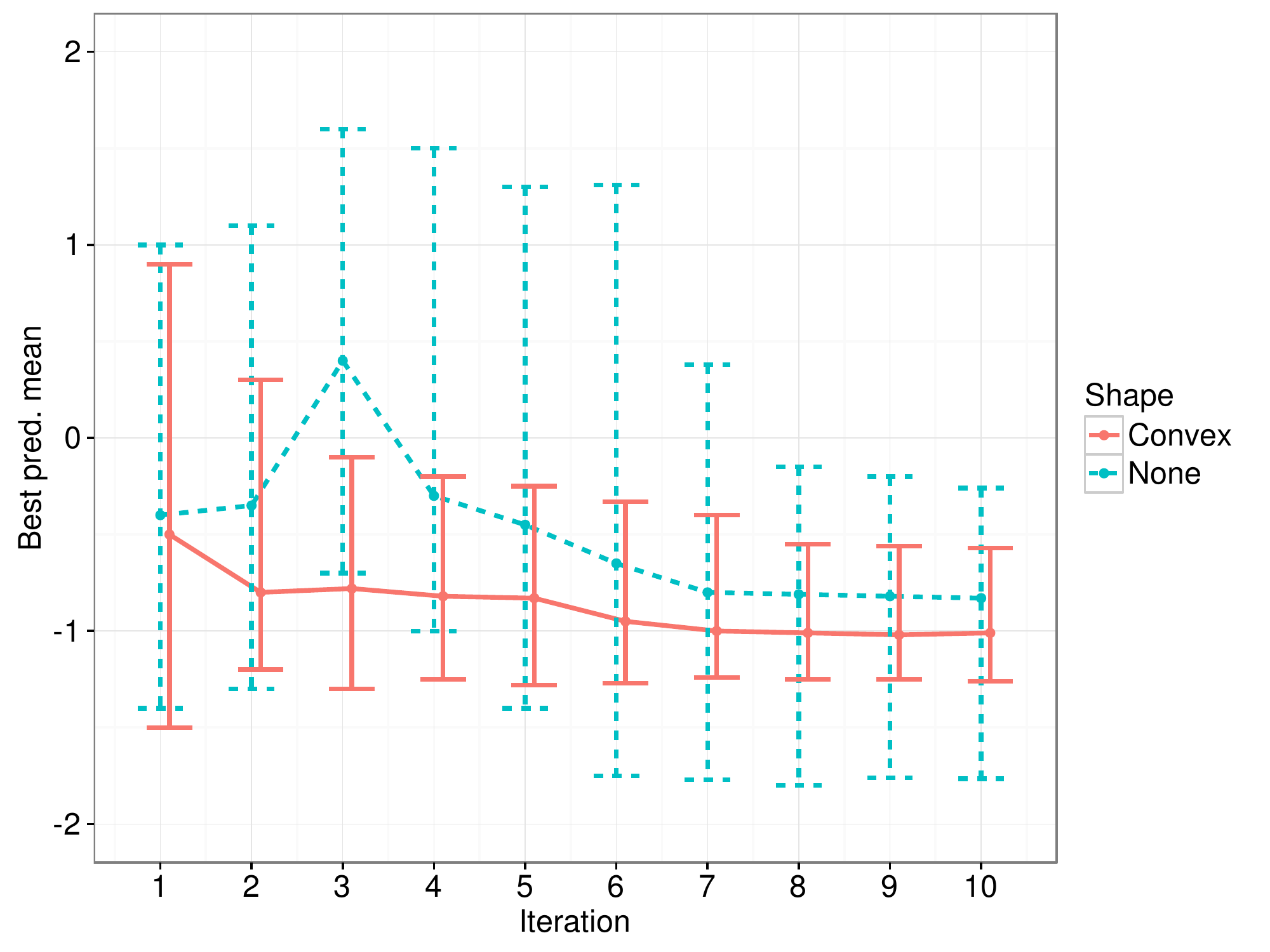}\label{fig:f2}}
  \caption{(a) simulated losses for sample sizes ranging from 1 to 120 (axes are standardized). (b) minimum posterior expected loss among the observed data, 50 random starting values.}
\label{fig:bin}
\end{figure}

This experiment is Example 1 in \citet{muller}. The goal is selecting the sample size of a binomial experiment where the prior on the probability of success $\theta$ is an equal-weighted mixture of a Beta(3,1) and a Beta(3,3). The loss of an experiment with sample size $n$ given data $y$ and $\theta$ is $L(n,y,\theta) = |\theta - m_y| + 0.0008n$ where $m_y$ is the posterior median. For any given $n$, the loss is estimated as the Monte Carlo average of 100 simulations from the joint distribution of $y$ and $\theta$.  The shape-constrained GP outperforms the unconstrained GP on average and appears to be more robust to the choice of starting points (see Figure~\ref{fig:bin}). In contrast to many Bayesian optimization papers, we plot smallest expected loss rather smallest function evaluation because the target is noisy. 
%\begin{figure}[h]
%  \centering
%  \fbox{\rule[-.5cm]{0cm}{4cm} \rule[-.5cm]{4cm}{0cm}}
%  \caption{Sample figure caption.}
%\end{figure}

\subsection{Support vector machine} \label{svm}

\begin{figure}[!htbp]
  \centering
  \subfloat[]{\includegraphics[width=0.49\textwidth]{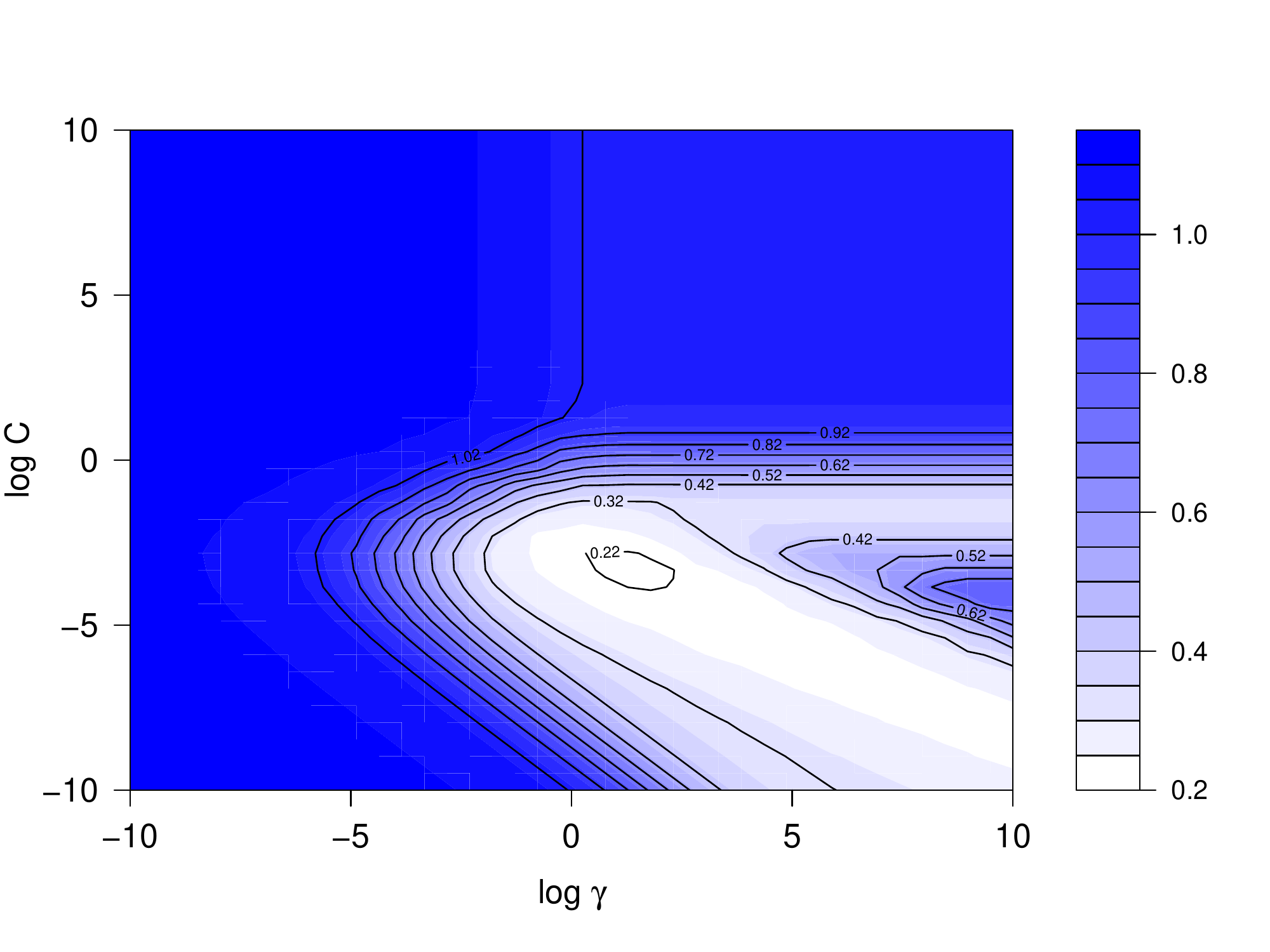}\label{fig:f3}}
  \hfill
  \subfloat[]{\includegraphics[width=0.49\textwidth]{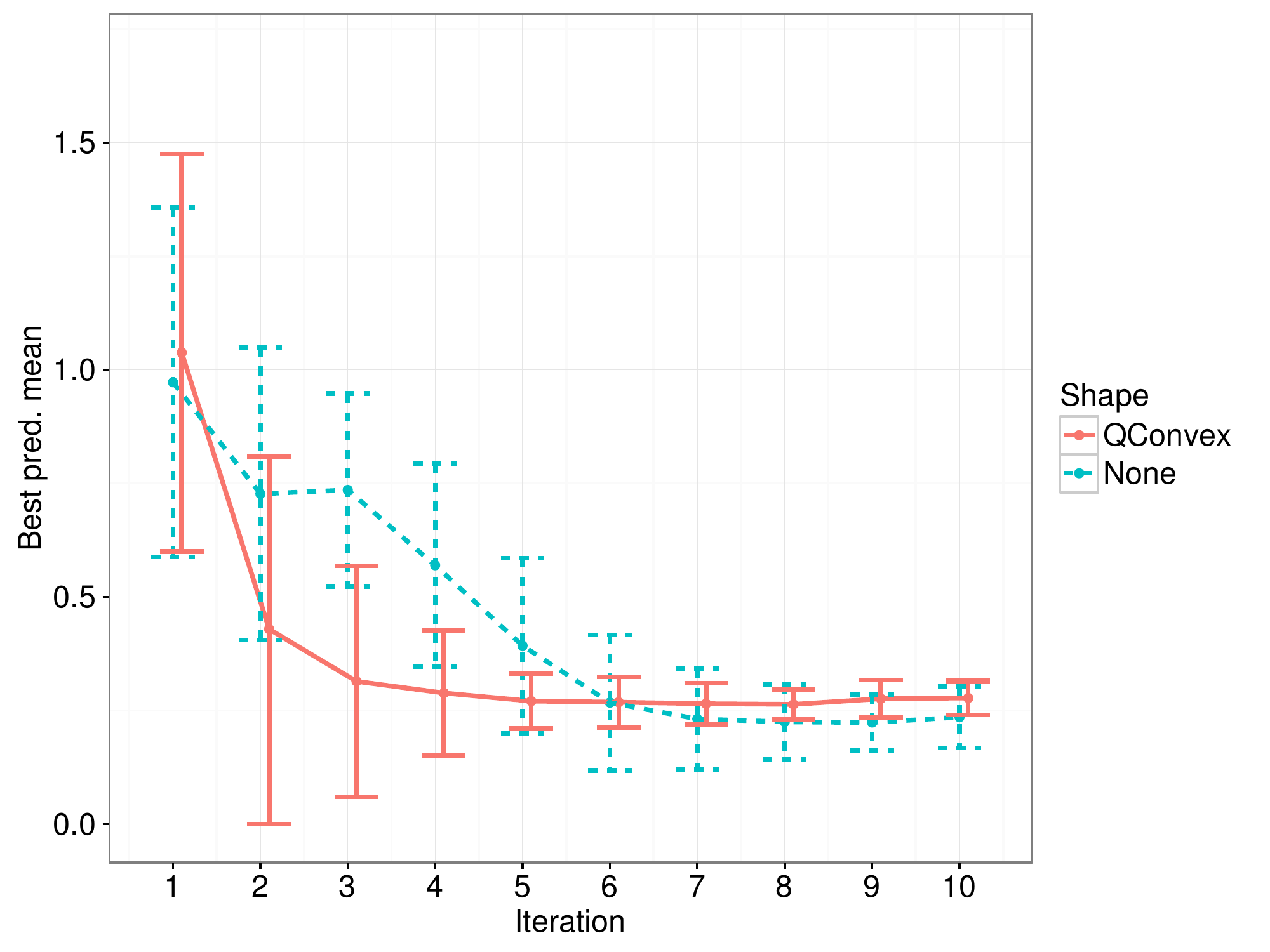}\label{fig:f4}}
  \caption{(a) ``true'' error surface. (b) 95\% CI posterior expected CV error.}
\label{fig:2d}
\end{figure}

The task is training an SVM with a squared-exponential kernel on the Ozone dataset (available in the \textsf{R} package \texttt{mlbench} \cite{mlbench}). We predict ozone readings using 12 other variables. We restrict $C , \gamma \in \{ \exp(x) : x \in [-10,10] \}$ and compare the performance of an unconstrained GP with that of a GP with quasiconvex constraints on both parameters. The performance metric is 10-fold cross-validated (CV) error. In this case, we only consider one starting point and compare the uncertainty in the minimum expected CV error in the observed samples. Figure~\ref{fig:2d} shows that the shape constrained GP stabilizes more quickly to a solution with lower uncertainty than the unconstrained version.

%\begin{figure}[h]
%  \centering
%  \fbox{\rule[-.5cm]{0cm}{4cm} \rule[-.5cm]{4cm}{0cm}}
%  \caption{Sample figure caption.}
%\end{figure}

%\section{Conclusion}

%We have presented an approach for ... and demonstrated benefits for simple problems. More work remains in extending the approach to more complex problems and many different choices could be make (other approaches to shape constraints, different covariance functions, different acquisition functions, etc.) Also, can be extended to other problem domains, such as decision analysis. Exciting stuff.

\newpage

\small

\bibliographystyle{plainnat}
\bibliography{references}

\iffalse
[1] Alexander, J.A.\ \& Mozer, M.C.\ (1995) Template-based algorithms
for connectionist rule extraction. In G.\ Tesauro, D.S.\ Touretzky and
T.K.\ Leen (eds.), {\it Advances in Neural Information Processing
  Systems 7}, pp.\ 609--616. Cambridge, MA: MIT Press.

[2] Bower, J.M.\ \& Beeman, D.\ (1995) {\it The Book of GENESIS:
  Exploring Realistic Neural Models with the GEneral NEural SImulation
  System.}  New York: TELOS/Springer--Verlag.

[3] Hasselmo, M.E., Schnell, E.\ \& Barkai, E.\ (1995) Dynamics of
learning and recall at excitatory recurrent synapses and cholinergic
modulation in rat hippocampal region CA3. {\it Journal of
  Neuroscience} {\bf 15}(7):5249-5262.
  
\fi 

\end{document}